\documentclass[10pt,twocolumn,letterpaper]{article}

\usepackage{times}
\usepackage{epsfig}
\usepackage{graphicx}
\usepackage{amsmath}
\usepackage{amssymb}
\usepackage{booktabs}

\usepackage{bm} %
\usepackage{xstring}
\usepackage{textcomp} %
\usepackage{float}
\usepackage{ifthen}
\usepackage[subrefformat=parens, labelformat=parens]{subcaption} %
\usepackage{gensymb}
\usepackage{tikz}
\usetikzlibrary{arrows,shapes,automata,petri,positioning,calc,decorations.pathreplacing}
\usepackage{pgfplots} %
\usepackage{xspace}
\usepackage{adjustbox}
\usepackage{multirow}
\usepackage{pifont}%
\usepackage{mfirstuc}
\pgfplotsset{compat=1.3}

\usepackage{color}      %

\newcommand{\tr}{\color{black}}
\newcommand{\tb}{\color{black}}

\newcommand{\myparagraph}{\paragraph}

\newcommand{\mabr}[1]{%
    \mbox{\textit{%
    \def\mabrplus{}%
    \let\mabrplus\undefined%
    \IfSubStr{#1}{v}{VIEW\def\mabrplus{}}{}%
    \IfSubStr{#1}{g}{\ifdef{\mabrplus}{+}{}GEO\def\mabrplus{}}{}%
    \IfSubStr{#1}{d}{\ifdef{\mabrplus}{+}{}d}{}%
    \IfSubStr{#1}{ll}{ (lin)}{}%
    \IfSubStr{#1}{nl}{ (MLP)}{}%
    }}%
}

\newcommand{\myshared}{redundant}
\newcommand{\myprivate}{complementary}

\newcommand{\mycmodal}{cross-modal}
\newcommand{\mycview}{cross-view}
\newcommand{\dataset}[1]{%
    \mbox{\textit{%
    \IfSubStr{#1}{nyuv2}{NYUv2}{}%
    \IfSubStr{#1}{scannet}{%
        \IfSubStr{#1}{scannet25k}{ScanNet25kframes}{ScanNet}%
    }{}%
    \IfSubStr{#1}{kitti}{KITTI}{}%
    \IfSubStr{#1}{cityscapes}{Cityscapes}{}%
    \IfSubStr{#1}{coco}{COCO}{}%
    }}%
}

\usepackage[pagebackref=true,breaklinks=true,colorlinks,bookmarks=false]{hyperref}

\usepackage[capitalize]{cleveref}
\Crefname{section}{Section}{Sections}
\crefname{section}{Section}{Sections}
\Crefname{table}{Table}{Tables}
\crefname{table}{Table}{Tables}
\crefname{figure}{Figure}{Figures}
\crefname{figure}{Figure}{Figures}

\begin{document}

\title{How do Cross-View and Cross-Modal Alignment\\ Affect Representations in Contrastive Learning?}

\author{Thomas M.\ Hehn 
\and
Julian F.P.\ Kooij \\[0.5em]
Intelligent Vehicles Group \\
TU Delft, The Netherlands
\and
Dariu M.\ Gavrila
}

\maketitle
\thispagestyle{empty}

\begin{abstract}
\tr{}
Various state-of-the-art self-supervised visual representation learning approaches take advantage of data from multiple sensors by aligning the feature representations across views and/or modalities.
In this work, we investigate how aligning representations affects the visual features obtained from \mycview{} and \mycmodal{} contrastive learning on images and point clouds.

On five real-world datasets and on five tasks, we train and evaluate 108 models based on four pretraining variations.
We find that \mycmodal{} representation alignment discards \myprivate{} visual information, such as color and texture, and instead emphasizes \myshared{} depth cues.
The depth cues obtained from pretraining improve downstream depth prediction performance.
Also overall, \mycmodal{} alignment leads to more robust encoders than pretraining by \mycview{} alignment, especially on depth prediction, instance segmentation, and object detection.
\tb{}

\end{abstract}

\section{Introduction}

\begin{figure}[t]
    \centering
    \subfloat[Intuition of \myprivate{} and \myshared{} information]{
        \resizebox{0.95\linewidth}{!}{\begin{tikzpicture}
    \def\firstcircle{(0,0) ellipse (3cm and 1.2cm)}
    \def\secondcircle{(0:3cm) ellipse (3cm and 1.2cm)}
    
    \tikzstyle{venncircle}=[white, opacity=1.0, font=\large];%
    \filldraw[olive, opacity=0.5] \firstcircle node[venncircle, left=0.2cm] {Visual (2D)};
    \filldraw[red, opacity=0.5] \secondcircle node[venncircle, right=0.0cm] {Geometric (3D)};
    
    \tikzstyle{feature}=[black, opacity=0.7];
    \node[feature] at (-1.0,0.8) {color};
    \node[feature] at (-0.6,-0.7) {texture};

    \node[feature] at (1.6,0.30) {depth};
    \node[feature] at (1.5,-0.30) {surfaces};
    
\end{tikzpicture}}}
    \\[1em]
    \subfloat[ScanNet image*\label{fig:pri3d_rgb}]{\includegraphics[width=0.45\linewidth]{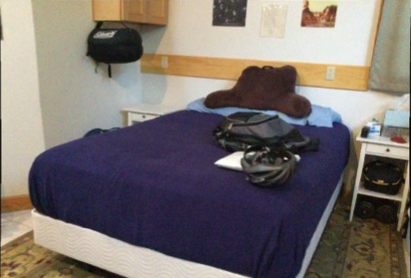}}
    ~
    \subfloat[ScanNet point cloud*\label{fig:pri3d_pc}]{\includegraphics[width=0.45\linewidth]{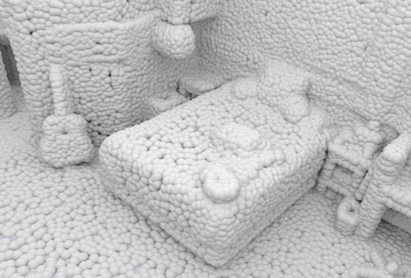}}
    \caption{\expandafter\makefirstuc\myshared{} information is shared across modalities, while \myprivate{} information is exclusive to a modality. In this work, we show that \mycmodal{} \textit{visual} representation learning predominantly encodes \myshared{} information of 2D and 3D data. *Images \subref{fig:pri3d_rgb} and \subref{fig:pri3d_pc} were taken from~\cite{hou2021pri3d}.}
    \label{fig:sharedprivate}
\end{figure}
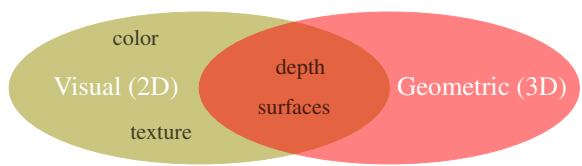
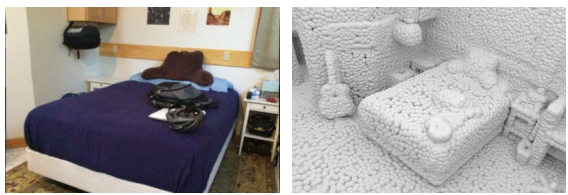

Pretraining of neural networks has been an established tool in computer vision for several years \cite{zeiler2014visualizing}.
It allows to successfully finetune neural networks on new tasks with fewer iterations and less annotated data~\cite{he2019rethinking}.
Commonly, models are initialized using weights pretrained for image classification on the ImageNet dataset~\cite{russakovsky15imagenet}, but recently, self-supervised approaches that do not rely on manual annotations have outperformed ImageNet pretraining~\cite{he2020momentum}.

In the realm of self-supervised representation learning, contrastive learning~\cite{Chen20SimCLR} has become a popular approach for visual representation learning.
In contrastive learning, the model learns to distinguish different variations of the same instance (e.g.~crops of a single image) from all other instances.
These variations can be created artificially or by using natural data correspondences, such as images and text~\cite{yuan2021multimodal} or from multiple viewpoints~\cite{tian2020contrastive,hou2021pri3d}.

A common objective is to enforce similarity between features across views or sensing modalities.
This is called \textit{representation alignment}~\cite{trosten2021reconsidering}.
While severe failure cases of such representation alignment have been discussed~\cite{trosten2021reconsidering,jaritz2020xmuda}, the effects on the representation itself were not studied, as of yet.

\Cref{fig:sharedprivate} shows a conceptual illustration of the information available in 2D image data and 3D point clouds in the ScanNet dataset~\cite{dai2017scannet}.
Some information, such as about depth and surfaces, is available in both modalities.
Although this information may be more detailed and more accurate in one modality than in the other, both modalities can express such properties to a certain degree.
In the context of sensor fusion, this is typically called \textit{\myshared{}} information.
Other information is exclusive to a modality, which is called \textit{\myprivate{}} information.
3D point clouds, for example, are generally color- and textureless.

Intuitively, a representation which discards \myprivate{} information is less expressive than one that includes \myshared{} and \myprivate{} information.
On the one hand, discarding texture and color information, for example, leads to a less complete visual representation.
On the other hand, ImageNet pretrained models were found to be texture-biased and by increasing their shape-bias the performance on tasks such as object detection could be improved~\cite{geirhos2019}.
So what is the influence of \myprivate{} and \myshared{} information on the visual representation quality?

In this work, we study how \myprivate{} and \myshared{} information affect visual representation learning when exploiting \mycview{} and \mycmodal{} representation alignment,
and how this relates to the performance of a finetuned model on a transfer learning downstream task.
For this purpose, we use Pri3D~\cite{hou2021pri3d}, a method that uses 3D information of a scene to enable \mycview{} and \mycmodal{} representation alignment for visual representations learning.

\section{Related work}

Self-supervised representation learning has recently surpassed supervised pretraining on ImageNet~\cite{he2020momentum}.
Various approaches have been explored to learn a global feature vector per image, such as contrastive learning~\cite{he2020momentum,Chen20SimCLR} and methods that do not require negative samples during training~\cite{chen2021exploring,grill2020byol}.
Since many computer vision tasks require per-pixel features, several works focused on learning dense representations~\cite{wang2020DenseCL,xie2021propagate}.
Inspired by this success for visual representations, contrastive learning has also been applied to 3D point cloud data~\cite{hou2021csc,xie2020pointcontrast,zhang2021self}.

Several approaches have explored cross-modal and cross-view visual representation learning.
While some propose a single model~\cite{liu2021contrastive} for multiple modalities, we are interested in training a vision-only model using cross-modal and cross-view input.
To this end, some have matched images and natural language to improve the learned visual representation~\cite{yuan2021multimodal,radford2021learning}, whereas others use 3D signals~\cite{li2022simipu,jing2021self,hou2021pri3d}, or match representations of different views of a scene~\cite{tian2020contrastive,hou2021pri3d}.

While \mycmodal{} feature spaces are a common technique~\cite{sayed2018cross,gupta2016cross,benton2017deep,hou2021pri3d,lee2021icra,zhang2021self},
they can introduce several caveats.
Especially when dealing with partially incomplete or corrupted data, representation alignment may hinder effective learning for multi-view clustering~\cite{trosten2021reconsidering}.
In domain adaptation, avoiding representation alignment across modalities has also proven to be beneficial to prevent discarding \myprivate{} sensor information~\cite{jaritz2020xmuda}. 
In representation learning, cross-modal training can even lead to an emphasis on a single strong modality and potentially harm the overall performance~\cite{liu2020p4contrast}.

Our goal is to understand the effects of \mycview{} and \mycmodal{} representation alignment on contrastive learning in more detail.
Specifically, we do not consider corrupted or incomplete sensor data, but we investigate how \myshared{} and \myprivate{} information influence the learned representations, and whether similar effects as observed for shape-biased vs texture-biased networks~\cite{geirhos2019} can be observed.
We base our empirical study on Pri3D~\cite{hou2021pri3d} as it uses both \mycview{} and \mycmodal{} representation alignment for contrastive learning, as well as their combination.
As our main contributions, we:
\begin{enumerate}
    \item assess how \mycview{} and \mycmodal{} representation alignment affect the \myprivate{} and \myshared{} information encoded in the learned visual per-pixel representations,
    \item evaluate the transfer learning performance in different settings on various tasks and datasets to investigate the robustness of \mycview{} and \mycmodal{} alignment.
\end{enumerate}

\section{Methods}
\label{sec:methods}

2D images and 3D point clouds have low-level \myshared{} and \myprivate{} features, as illustrated in \cref{fig:sharedprivate}.
Therefore, these modalities lend themselves to study \mycview{} and \mycmodal{} representation alignment.
Our analysis of alignment in representation learning is based on Pri3D \cite{hou2021pri3d}.
Pri3D uses 3D information for 2D visual representation learning.
It employs a self-supervised contrastive learning approach that does not require any human annotation, but only RGB images and registered point clouds that can be obtained, for example, from an RGB-D dataset.
This section summarizes the two contrastive losses of Pri3D and discusses how the feature spaces of the losses can be separated.%

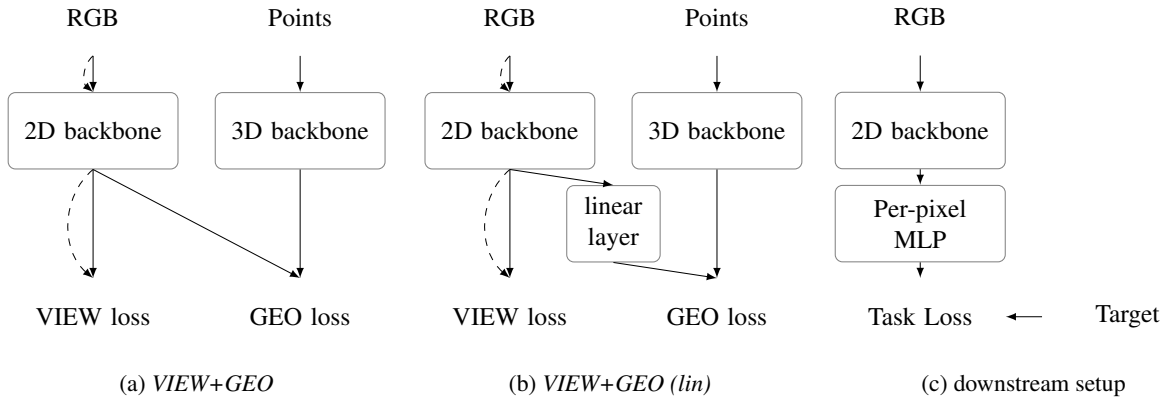
\begin{figure*}[tb]
    \centering
    \subfloat[\mabr{vgno}\label{fig:vgno}]{
        \resizebox{0.32\linewidth}{!}{
            \begin{tikzpicture}
    \providecommand{\showcodegraph}{false}
    \providecommand{\showdepthgraph}{false}
    \def\nodewidth{6.0em}
    \def\nodeheight{3.0em}
    \def\nodedistx{0.5}
    \tikzstyle{mainnode}=[black!50, text=black,  text centered, minimum height=\nodeheight, text width=\nodewidth]
    \tikzstyle{fixednode}=[fill=gray!25]
    \tikzstyle{NNnode}=[draw, rounded corners=0.1cm, text width=\nodewidth]
    \tikzset{>=latex}
    
    \tikzset{
    between/.style args={#1 and #2}{
         at = ($(#1)!0.5!(#2)$)
        }
    }
    
    \node[mainnode] (input_rgb) {RGB};
    \node[mainnode, right=\nodedistx of input_rgb] (input_pc) {Points};
    \node[mainnode, below=\nodedistx of input_rgb, NNnode]  (backbone2d) {2D backbone};
    \node[mainnode, below=\nodedistx of input_pc, NNnode]  (backbone3d) {3D backbone};
    
    \node[mainnode, below=(3*\nodedistx) of backbone2d]  (viewloss) {VIEW loss};
    \node[mainnode, below=(3*\nodedistx) of backbone3d]  (geoloss) {GEO loss};
    
    \path[draw,->] (input_rgb.south) -- (backbone2d.north);
    \path[draw,->] (input_pc.south) -- (backbone3d.north);
    \path[draw,->] (backbone2d.south) -- (viewloss.north);
    
    \path[->, dashed] (input_rgb.south) edge [bend right=50] (backbone2d.north);
    \path[->, dashed] (backbone2d.south) edge [bend right=50] (viewloss.north);
    
    \ifthenelse{\equal{\showcodegraph}{false}}{
        \path[draw, ->] (backbone2d.south) -- (geoloss.north);
    }{
        \node[mainnode, between=backbone2d and geoloss, NNnode, text width=(0.5 * \nodewidth)]  (w2d) {\showcodegraph};
        \path[draw, ->] (backbone2d.south) -- (w2d.north);
        \path[draw, ->] (w2d.south) -- (geoloss.north);
    }
    \ifthenelse{\equal{\showdepthgraph}{true}}{
        \node[mainnode, left=\nodedistx of viewloss]  (depthloss) {Depth loss};
        \node[mainnode, NNnode, text width=(0.5 * \nodewidth), between=backbone2d and depthloss]  (headnet) {linear\\layer};
        \path[draw, ->] (backbone2d.south) -- (headnet.north);
        \path[draw, ->] (headnet.south) -- (depthloss.north);
    }{ }
    \path[draw, ->] (backbone3d.south) -- (geoloss.north);
\end{tikzpicture}
        }
        \let\showcodegraph\undefined
    }
    \subfloat[\mabr{vgll}\label{fig:vgll}]{
        \newcommand{\showcodegraph}{linear layer} 
        \resizebox{0.32\linewidth}{!}{
        \begin{tikzpicture}
    \providecommand{\showcodegraph}{false}
    \providecommand{\showdepthgraph}{false}
    \def\nodewidth{6.0em}
    \def\nodeheight{3.0em}
    \def\nodedistx{0.5}
    \tikzstyle{mainnode}=[black!50, text=black,  text centered, minimum height=\nodeheight, text width=\nodewidth]
    \tikzstyle{fixednode}=[fill=gray!25]
    \tikzstyle{NNnode}=[draw, rounded corners=0.1cm, text width=\nodewidth]
    \tikzset{>=latex}
    
    \tikzset{
    between/.style args={#1 and #2}{
         at = ($(#1)!0.5!(#2)$)
        }
    }
    
    \node[mainnode] (input_rgb) {RGB};
    \node[mainnode, right=\nodedistx of input_rgb] (input_pc) {Points};
    \node[mainnode, below=\nodedistx of input_rgb, NNnode]  (backbone2d) {2D backbone};
    \node[mainnode, below=\nodedistx of input_pc, NNnode]  (backbone3d) {3D backbone};
    
    \node[mainnode, below=(3*\nodedistx) of backbone2d]  (viewloss) {VIEW loss};
    \node[mainnode, below=(3*\nodedistx) of backbone3d]  (geoloss) {GEO loss};
    
    \path[draw,->] (input_rgb.south) -- (backbone2d.north);
    \path[draw,->] (input_pc.south) -- (backbone3d.north);
    \path[draw,->] (backbone2d.south) -- (viewloss.north);
    
    \path[->, dashed] (input_rgb.south) edge [bend right=50] (backbone2d.north);
    \path[->, dashed] (backbone2d.south) edge [bend right=50] (viewloss.north);
    
    \ifthenelse{\equal{\showcodegraph}{false}}{
        \path[draw, ->] (backbone2d.south) -- (geoloss.north);
    }{
        \node[mainnode, between=backbone2d and geoloss, NNnode, text width=(0.5 * \nodewidth)]  (w2d) {\showcodegraph};
        \path[draw, ->] (backbone2d.south) -- (w2d.north);
        \path[draw, ->] (w2d.south) -- (geoloss.north);
    }
    \ifthenelse{\equal{\showdepthgraph}{true}}{
        \node[mainnode, left=\nodedistx of viewloss]  (depthloss) {Depth loss};
        \node[mainnode, NNnode, text width=(0.5 * \nodewidth), between=backbone2d and depthloss]  (headnet) {linear\\layer};
        \path[draw, ->] (backbone2d.south) -- (headnet.north);
        \path[draw, ->] (headnet.south) -- (depthloss.north);
    }{ }
    \path[draw, ->] (backbone3d.south) -- (geoloss.north);
\end{tikzpicture}
        }
    }
    \subfloat[downstream setup\label{fig:downstreamsetup}]{
        \resizebox{0.32\linewidth}{!}{
        \begin{tikzpicture}
    \def\nodewidth{6.0em}
    \def\nodeheight{3.0em}
    \def\nodedistx{0.5}
    \tikzstyle{mainnode}=[black!50, text=black,  text centered, minimum height=\nodeheight, text width=\nodewidth]
    \tikzstyle{fixednode}=[fill=gray!25]
    \tikzstyle{NNnode}=[draw, rounded corners=0.1cm, text width=\nodewidth]
    \tikzset{>=latex}

    \tikzset{
    between/.style args={#1 and #2}{
         at = ($(#1)!0.5!(#2)$)
        }
    }
    
    \providecommand{\showsmallgraph}{true}
    \ifthenelse{\equal{\showsmallgraph}{true}}{
        \node[mainnode] (input) {RGB};
        \node[mainnode, below=\nodedistx of input, NNnode]  (backbone) {2D backbone};
        \node[mainnode, below=(3*\nodedistx) of backbone]  (loss) {Task Loss};
        \node[mainnode, between=backbone and loss, NNnode]  (MLP) {Per-pixel MLP};
        \node[mainnode, right=\nodedistx of loss]  (groundtruth) {Target};
        
        \path[draw, ->] (input.south) -- (backbone.north);
        \path[draw, ->] (backbone.south) -- (MLP.north);
        \path[draw, ->] (MLP.south) -- (loss.north);
        \path[draw, ->] (groundtruth.west) -- (loss.east);
    }{ %
        \node[mainnode] (input) {RGB};
        \node[mainnode, below=\nodedistx of input, NNnode, fixednode]  (backbone) {2D backbone};
        \node[mainnode, below=\nodedistx of backbone, NNnode]  (1x1_1) {1x1@128Conv. + BN + ReLU};
        \node[mainnode, below=\nodedistx of 1x1_1, NNnode]  (1x1_2) {1x1@128Conv. + BN + ReLU};
        \node[mainnode, below=\nodedistx of 1x1_2, NNnode]  (1x1_3) {1x1@(3/1)Conv.};
        \node[mainnode, below=\nodedistx of 1x1_3]  (interp) {Bilinear upscaling (2x)};
        \node[mainnode, below=\nodedistx of interp]  (loss) {L2 Loss};
        \node[mainnode, left=\nodedistx of loss]  (groundtruth) {RGB/Depth ground truth};
        
        \path[draw, ->] (input.south) -- (backbone.north);
        \path[draw, ->] (backbone.south) -- (1x1_1.north);
        \path[draw, ->] (1x1_1.south) -- (1x1_2.north);
        \path[draw, ->] (1x1_2.south) -- (1x1_3.north);
        \path[draw, ->] (1x1_3.south) -- (interp.north);
        \path[draw, ->] (interp.south) -- (loss.north);
        \path[draw, ->] (groundtruth.east) -- (loss.west);
    }
\end{tikzpicture}
        }
    }
    \caption{%
    During pretraining, the combination \protect\mabr{vgno} \subref{fig:vgno} optimizes the same features in the \protect\mabr{v} and \protect\mabr{g} loss, while the linear layer in \protect\mabr{vgll} \subref{fig:vgll} separates the feature spaces of the losses.
    The dashed arrows indicate a second overlapping view.
    Only the 2D backbone performance is evaluated \subref{fig:downstreamsetup}.}
    \label{fig:pri3dpapervscode}
\end{figure*}

\subsection{Pri3D: contrastive losses}
Pri3D finds pairs of pixels and elements of a regular grid in 3D space, so-called voxels, for contrastive learning.
Pixels and voxels are matched by their distance in 3D space to define two losses.
One loss requires pairs of pixels across views, whereas the other loss is based on pairs of pixels and 3D points across modalities.
Since Pri3D matches pixels and voxels based on their 3D positions, it is relatively robust with respect to incomplete sensor data.

\myparagraph{\expandafter\makefirstuc\mycview{} contrastive loss (\protect\mabr{v})}
This contrastive loss matches pixels across different views of the same scene.
It requires two RGB-D images from the same scene and the corresponding relative camera poses and calibration.
Once two pixels from two different views \textit{A} and \textit{B} are within a 2 cm distance, they are considered a positive pair in the sense of contrastive learning~\cite{Chen20SimCLR}.
Only pixels that are part of a positive pair are used in this loss, as the negative pairs are also constructed from these pixels.
To form the negative pairs, each pixel $a$ of view \textit{A} of a positive pair $(a,b)$ is paired with all pixels $k$ of view \textit{B} which are part of a positive pair but not $k = b$.

For each image $x$, an encoder-decoder network $f$ predicts a set feature vectors $f(x)$, and $\bm{f}_a$ denotes the L2-normalized feature vector of pixel $a$.
Using the set of positive pairs $M$, Pri3D employs a PointInfoNCE loss \cite{xie2020pointcontrast}
\begin{equation}
    \mathcal{L}_p = - \sum_{(a,b) \in M} \log \frac{\exp(\bm{f}_a \cdot \bm{f}_b / \tau)}{\sum_{(\cdot,k) \in M} \exp(\bm{f}_a \cdot \bm{f}_k / \tau)},
    \label{eq:pointinfonce}
\end{equation}
where $\tau$ is a temperature hyperparameter.
This \mycview{} loss is called the \mabr{v} loss, following the convention of Pri3D.
Compared to other visual representation learning approaches, \mabr{v} does not rely on strong augmentations that may corrupt color information~\cite{Chen20SimCLR}.

\myparagraph{\expandafter\makefirstuc\mycmodal{} contrastive loss (\protect\mabr{g})}
Unlike the \mabr{v} loss, the pairs for the \mabr{g} loss do not consist of two pixels from different views, but instead of a pixel and a voxel.
The positive pairs are again pixels and voxels within 2 cm distance in 3D space.
The negative pairs are all other non-matching pixels and voxels pairs, similar to the \mabr{v} loss.

The voxel is associated with a feature vector obtained from a dense 3D feature encoder.
This dense 3D feature encoder is jointly learned with the dense 2D feature encoder.
The loss function (\cref{eq:pointinfonce}) remains the same with $\bm{f}_a$ and $\bm{f}_b$ now being feature vectors from the 2D and 3D feature encoders, respectively.
Thus, this \mycmodal{} loss enforces the features to be similar across the two modalities and is called the \mabr{g} loss, following the convention of Pri3D.

\subsection{Pri3D: representation space separation}
In Pri3D~\cite{hou2021pri3d}, the \mabr{g} and the \mabr{v} loss use the same visual feature vector for a single pixel.
Thus, the combination \mabr{vg} enforces mutual alignment of \mycview{} \textit{and} \mycmodal{} representations.
Interestingly, in the implementation published by the original Pri3D authors\cite{hou2021pri3dcode}, the feature spaces of both losses are separated by a linear layer preceding the \mabr{g} loss (see \cref{fig:pri3dpapervscode}).
We hypothesize that the original purpose of this layer was merely to match the feature dimensions of the 2D and 3D backbone (e.g.~for ResNet50 2D backbone this layer projects 128 to 96 channels).
Yet, this separation of feature spaces relaxes the alignment of the \mycview{} with the \mycmodal{} feature representation.
The feature vector of a pixel in the \mabr{g} loss is now computed by a linear function on the feature vector of the same pixel in the \mabr{v} loss, thus the two feature vectors are not the same anymore.
The \mycview{} and the \mycmodal{} representations are not directly aligned.
Since the effect of this linear layer was not described in the literature, we include both variants in our experiments.

In the following, we will distinguish this variation by adding\mabr{ll} to models that include the linear layer that separates the feature representations.
These and other differences to the original Pri3D paper are described in more detail in the supplementary material.

\section{Experiments}
\tr{}
In our experiments, we examine how \mycview{} and \mycmodal{} representation alignment in Pri3D influence the encoded \myprivate{} and \myshared{} information and how possible differences in the representations affect downstream transfer learning performance.

\subsection{Setup}
The experiments consist of two main steps: We first pretrain models based on the variations of Pri3D (see \cref{sec:methods}).
Afterward, the pretrained models are used to initialize training on downstream tasks, where we distinguish between a frozen (\cref{sec:frozendownstream}), a half-frozen (\cref{sec:exp_halffrozen}), and a full finetuning (\cref{sec:exp_fullfinetuning}) setting.
\tb{}

\subsubsection{Pretraining}
Our implementation is mainly based on the published code of Pri3D, and thus, we follow the same pretraining protocol as presented in the paper~\cite{hou2021pri3d} and give a coarse outline here.
We always use a UNet-style ResNet50 2D backbone and a UNet-style sparse convolutional 3D backbone.
As in~\cite{hou2021pri3d}, the 2D backbone is initialized using supervised ImageNet pretraining before performing self-supervised pretraining on the \dataset{scannet} dataset~\cite{dai2017scannet}.
The \dataset{scannet} dataset is a collection of RGB-D sequences of indoor scenes, captured with a Kinect-like setup.
On this dataset, the model is trained for 5 epochs using stochastic gradient descent with a learning rate of 0.1 with polynomial decay of 0.9 and the batch size is 64, unless explicitly mentioned otherwise.
This corresponds to approximately 60k iterations and 4 days of training time on 8 Nvidia V100 GPUs for pretraining a single model.
The temperature parameter in the losses (\cref{eq:pointinfonce}) is set to 0.4.

\subsubsection{Per-pixel downstream tasks}

In \cref{sec:frozendownstream,sec:exp_halffrozen,sec:exp_fullfinetuning} the per-pixel representations are evaluated on depth prediction, image reconstruction, and semantic segmentation while parts of the pretrained network are frozen.
To decode the information from features extracted by the frozen pretrained network, we append a non-linear per-pixel MLP (see \cref{fig:downstreamsetup}). %
This MLP is implemented as a sequence of 1x1 convolutions: %
Conv1x1@128 $\rightarrow$ ReLU $\rightarrow$ BatchNorm $\rightarrow$ Conv1x1@128 $\rightarrow$ ReLU $\rightarrow$
BatchNorm $\rightarrow$ Conv1x1@$c$ ($c=3$ for image reconstruction, $c=1$ for depth estimation, and $c=\#classes$ for semantic segmentation).
The MLP is followed by a bilinear fixed interpolation that upscales the output from $120 \times 160$ to $240 \times 320$, which is analogous to the semantic segmentation downstream experiments of Pri3D~\cite{hou2021pri3d} with the UNet architecture.
Note that we discard the last convolutional layer from the pretrained model, which, during pretraining, linearly projects the features to the space in which the contrastive loss operates. It can be regarded as the head network which aims to solve the contrastive task.

For image reconstruction and depth prediction, we choose an L2 loss directly on the RGB input pixel values and depth ground truth values, respectively.
For semantic segmentation, we use the common cross-entropy loss to predict the semantic classes of each pixel.
Other than that, the same training protocol as for the semantic segmentation task in \cite{hou2021pri3d} is applied.
The mean Intersection-over-Union (mIoU) is the metric for the semantic segmentation quality.
The datasets used for depth prediction and image reconstruction are:
\begin{itemize}
    \item \dataset{scannet25k}~\cite{dai2017scannet}: a subset of the \dataset{scannet} dataset annotated for specific tasks, e.g.~semantic segmentation, with given training/test/validation splits. The RGB input images of all three splits are subsets of the \dataset{scannet} dataset used for pretraining. On all downstream tasks, we always use \dataset{scannet25k} and therefore do not distinguish it from \dataset{scannet} explicitly.
    \item \dataset{nyuv2}~\cite{silberman2012nyuv2}: A dataset of 1449 RGBD images with semantic segmentation annotations, capturing 464 indoor scenes.
\end{itemize}
For semantic segmentation, we additionally use the following datasets:
\begin{itemize}
    \item \dataset{kitti}~\cite{alhaija2018kitti}: The dataset of the KITTI semantic segmentation benchmark consists of 200 semantically annotated training as well as 200 test images (our validation set) of road traffic scenes.
    \item \dataset{cityscapes}~\cite{cordts2016cityscapes}: A dataset of 5000 semantically annotated frames of urban street scenes in 50 different cities.
\end{itemize}

\subsubsection{Object-centered downstream tasks}
To perform object detection and instance segmentation, the encoders of the pretrained models are used to initialize a Mask RCNN model~\cite{he2017mask} of the Detectron2 library~\cite{wu2019detectron2} in \cref{sec:objecttasks}.
Due to the downstream architecture, the decoders are discarded.
In addition to \dataset{scannet25k} and \dataset{nyuv2}, the models are evaluated on the 2017 version of the \dataset{coco} dataset~\cite{lin2014coco} that contains more than 118k training and 5000 validation images.
The Average Precision (AP) metric is used on all datasets.

\tr{}
\begin{table*}[t!]
  \centering
  \scalebox{1.0}{%
  \begin{tikzpicture}
      \node[anchor=north west] (full) at (0,0) {\includegraphics[]{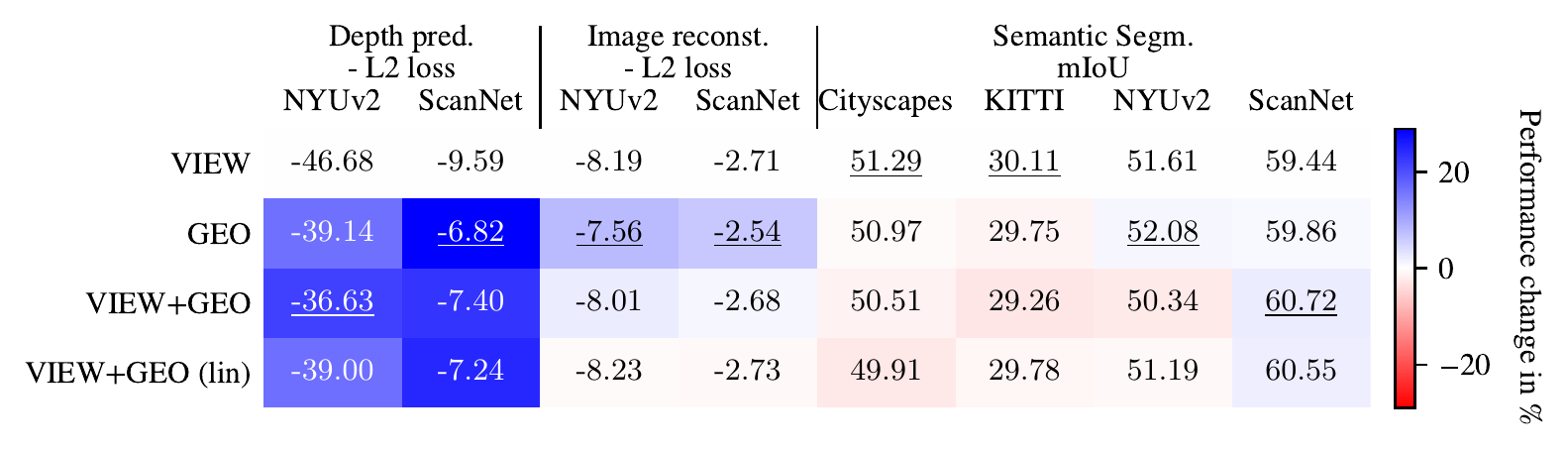}};
      \node[above=of full.center, anchor=south, yshift=0.1cm] (halffrozen) {\includegraphics[]{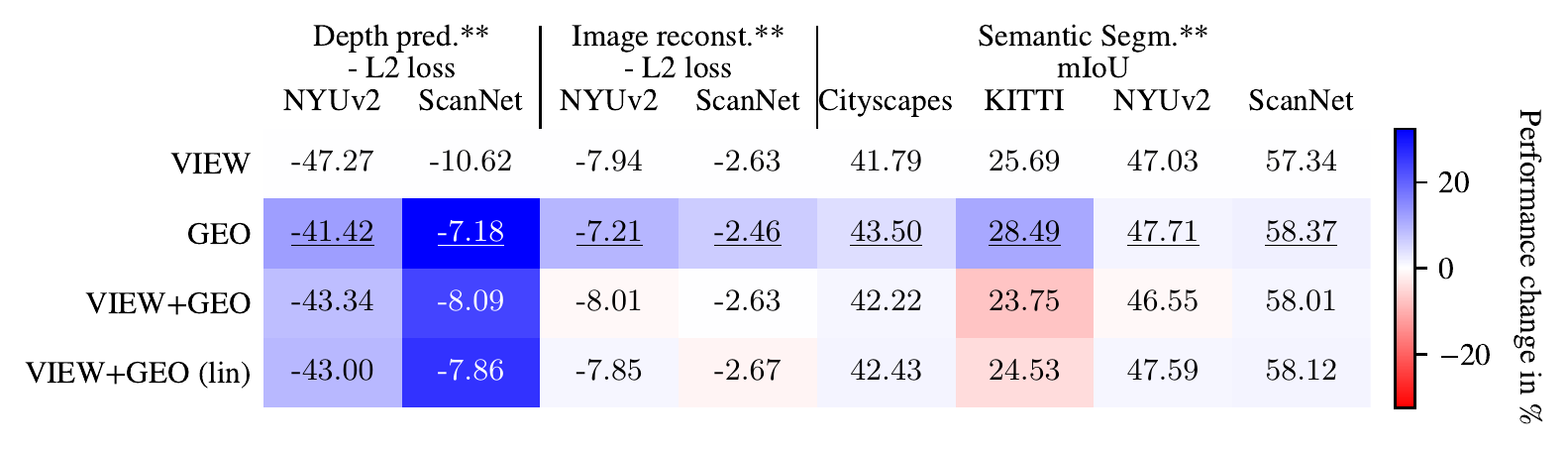}};
      \node[above=of halffrozen.center, anchor=south, yshift=0.1cm] (frozen) {\includegraphics[]{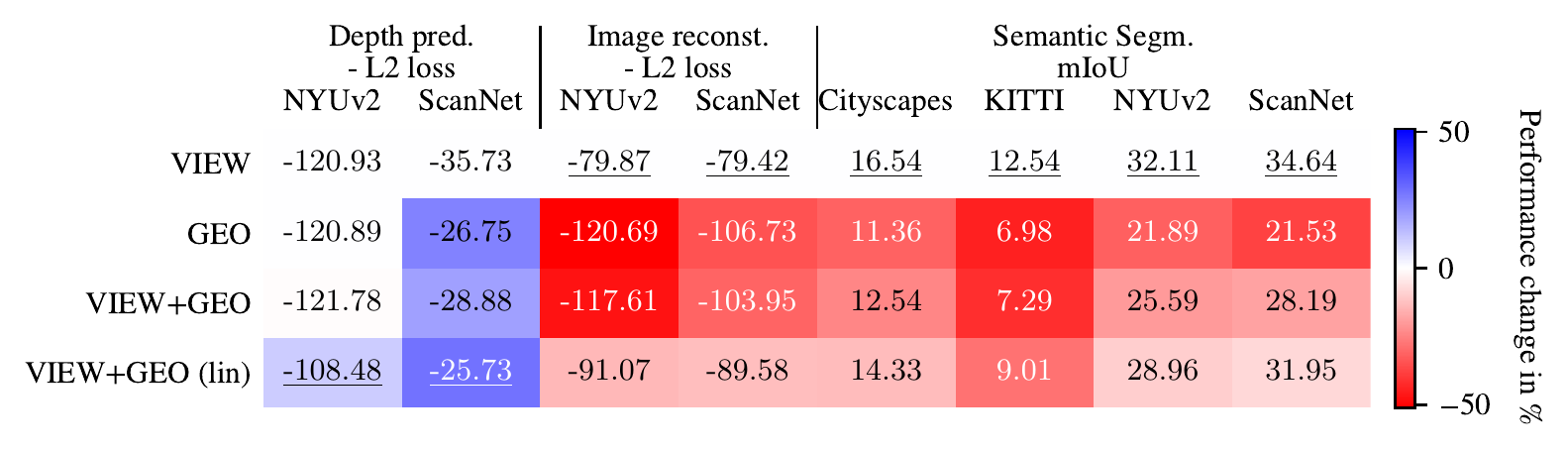}};
      \fill[black] (0,-1.15) rectangle ++(15,0.03);
      \fill[white] (2.5,-1.15) rectangle ++(11.7,-0.29);
      \fill[black] (0,2.35) rectangle ++(15,0.03);
      \fill[white] (2.5,2.35) rectangle ++(11.7,-0.29);
      \node[anchor=north, rotate=90] at (frozen.west) {Frozen};
      \node[anchor=north, rotate=90] at (halffrozen.west) {Half-frozen};
      \node[anchor=north, rotate=90] at (full.west) {Full finetuning};
  \end{tikzpicture}
  }
  \caption{%
        \tr{}%
           Validation performance on the three tasks and four datasets.
           Rows 1-4 show the frozen downstream setting where the pretrained weights are fixed, in rows 5-8 the encoder of the UNet-style architecture is fixed, and in the last four rows, the entire models are finetuned.
           To have a consistent notion that a greater value means better performance, the negative L2-Loss is shown.
           Blue color indicates relative improvement with respect to the \mycview{} model (\protect\mabr{v}). %
        \tb{}%
           }
  \label{tab:hm_ViewvGeovViewGeoFrozen}
\end{table*}

\subsection{Frozen downstream tasks}
\label{sec:frozendownstream}
With the frozen downstream tasks, where we keep all weights of the pretrained network frozen, we investigate how well the pretrained models capture color, depth, or semantic information by predicting this information from the learned per-pixel representations.
Using these frozen downstream tasks, we show how the different alignment strategies in Pri3D influence the learned representations.
For semantic segmentation, this frozen setup is similar to the linear evaluation protocol commonly used to evaluate the quality of global image features in visual representation learning~\cite{Chen20SimCLR}.
In this linear evaluation protocol, the performance of a linear classifier on the global features is often regarded as a proxy metric for the representation quality.
In our setup, we evaluate whether a non-linear MLP can provide a proxy metric for the representation quality.

\begin{figure*}[tb]
    \centering
    \input{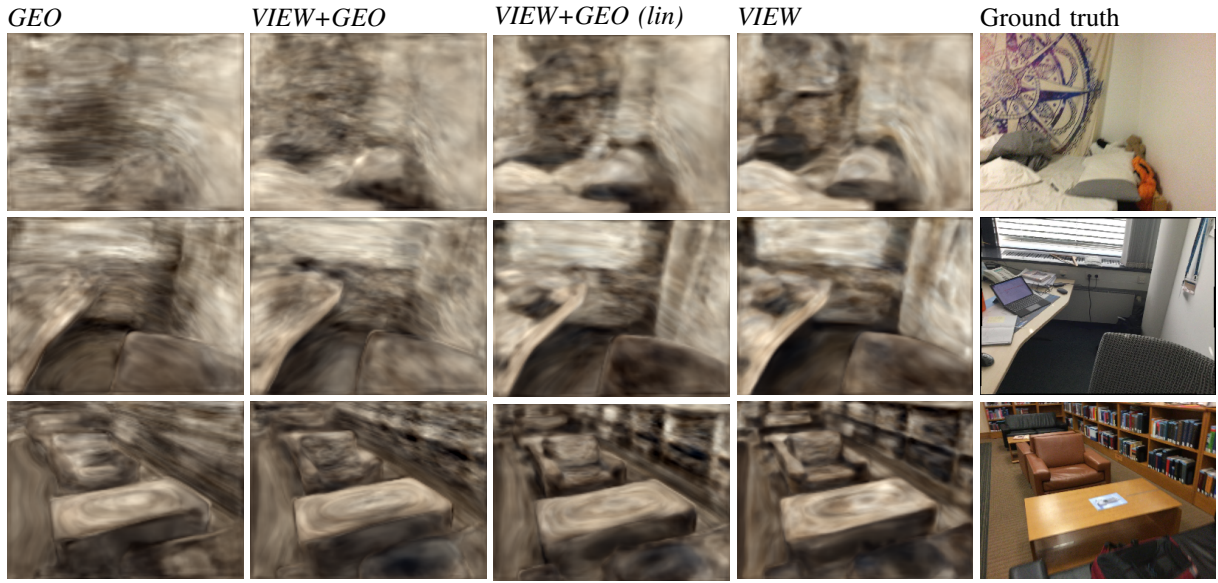}
    \caption{%
    Image reconstructions obtained from different frozen representations.
    The rows correspond to different samples.
    The rightmost column is the ground truth data and also the input for the feature extractors.
    The other columns use different models to compute the features from which the images are reconstructed.}
    \label{fig:qualitative_reconst}
\end{figure*}

\tb{}

\begin{figure*}[tb]
    \centering
    \input{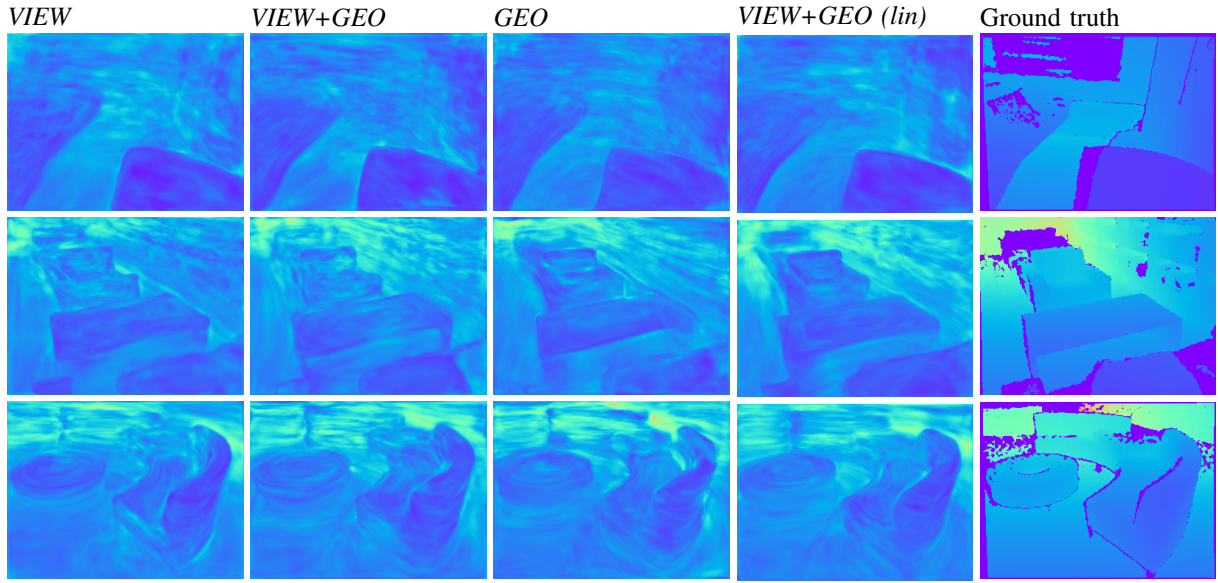}
    \caption{
    Qualitative results on the frozen depth prediction task.
    Each row corresponds to one sample and the rightmost column shows the ground truth depth data (saturated blue indicates invalid values).
    The other columns use different models to compute the frozen per-pixel features for depth prediction.%
    }
    \label{fig:qualitative_depth}
\end{figure*}

\tr{}

The first four rows of \cref{tab:hm_ViewvGeovViewGeoFrozen} show the best performance on the validation data during the training process for each model variation and frozen downstream task.
Compared to the \mycview{} model (\mabr{v}), all other models perform worse on frozen image reconstruction and semantic segmentation.
In contrast, the models that include the \mycmodal{} \mabr{g} loss perform better than the \mabr{v} on ScanNet depth prediction, while on NYUv2 this general trend is not visible.
The \mabr{vgll} model outperforms the \mabr{vgno} model on all three frozen downstream tasks by more than 10\%.

\Cref{fig:qualitative_reconst,fig:qualitative_depth} show the qualitative reconstruction and depth prediction results.  %
In both cases, the ranking of the models on ScanNet is reflected in the visual reconstruction quality.
The \mabr{v} model achieves the best visual reconstruction, followed by the \mabr{vgll}, \mabr{vgno} and \mabr{g} models.
It is notable that the results differ especially on flat texture-rich surfaces, for example in the first row where the \mabr{g} model does recover the wallpaper.
All model variations discard most of the color information.
The most detailed and smoothest depth estimation is obtained by the \mabr{vgll} model.
The depth gradient along surfaces appears smooth whereas the depth estimation based on \mabr{vgno} and \mabr{v} is speckled on flat surfaces.

In conclusion, the \mycview{} \mabr{v} loss preserves more texture information while the \mycmodal{} \mabr{g} loss leads to a better depth representation.
This observation is in line with the ideas of \myprivate{} and \myshared{} information discussed in \cref{fig:sharedprivate}.
Interestingly, combining the two losses as \mabr{vgno} and aligning the representations across views and modalities only achieves mediocre performance on all three downstream tasks.
Once alignment is lifted by the linear layer in \mabr{vgll}, all metrics improve, showing that it allows the model to preserve more \myprivate{} information.
Notably, the depth prediction result is even better than for the \mabr{g} model, suggesting that even with respect to the \myshared{} information, this separation is beneficial.
The semantic segmentation performance differs strongly between the model variations and benefits from the \mycview{} loss and the linear layer in \mabr{vgll} that lifts the direct alignment.

\subsection{Half-frozen downstream tasks}
\label{sec:exp_halffrozen}
By freezing only the encoder part of the UNet-style models, the following experiments provide insight to which part of architecture discards the texture information.
The experimental setup, architecture, and datasets remain the same as for the frozen downstream tasks in \cref{sec:frozendownstream}, except that now also the decoder is finetuned together with the MLP and only the encoder is fixed during finetuning.

The \mabr{g} model performs best across all tasks and datasets as seen in rows 5-8 of \cref{tab:hm_ViewvGeovViewGeoFrozen}.
The model excels on depth prediction, and interestingly, also outperforms the \mabr{v} model on image reconstruction.
The combined models outperform \mabr{v} on depth prediction, but the other tasks show mixed results.
The linear layer\mabr{ll} improves the \mabr{vg} performance slightly.

Overall, the results are in stark contrast to the frozen downstream tasks.
Given that the \mabr{g} model outperforms the \mabr{v} model, it seems that the decoder discards the texture information in the frozen downstream tasks, and recovering the texture is possible by finetuning.
At the same time, the encoder of the \mabr{g} model extracts very robust features across tasks and datasets.

\subsection{Full finetuning downstream tasks}
\label{sec:exp_fullfinetuning}

To assess the practical value and generality of a visual representation, the models are now fully finetuned and evaluated on the downstream tasks.
In contrast to the frozen downstream tasks, all weights in the model are finetuned during the downstream task training.%
\footnote{Note that the results deviate from the results presented in \cite{hou2021pri3d} since we discard its proxy depth loss (see supplementary material).}
As the full model is finetuned, the per-pixel MLP is not needed anymore and is omitted.

Rows 9-12 in \cref{tab:hm_ViewvGeovViewGeoFrozen} show the full finetuning downstream performance.
Models that were trained including the \mabr{g} loss outperform the \mabr{v} model on depth prediction.
Again, \mabr{g} is the best model for image reconstruction.
On semantic segmentation, \mabr{v} performs best on \dataset{cityscapes} and \dataset{kitti}, but not on the other two datasets.

Also in this full finetuning setting, \mycmodal{} alignment improves depth prediction performance.
For semantic segmentation, the results are generally mixed, but it is notable that \mabr{v} performs well on the two outdoor datasets.
This may indicate that the \mycmodal{} models are biased towards spatial layouts of indoor scenes.
Adding the feature space separation\mabr{ll} to \mabr{vg} does not have a positive effect, but rather tends to decrease the performance.

\subsection{Instance and object segmentation}
\label{sec:objecttasks}
As discussed in~\cite{geirhos2019}, texture-biased networks are inferior to shape-biased networks for object detection.
In order to test whether this also holds for texture-biased networks compared to depth-biased networks, the pretrained encoders are finetuned for object detection and instance segmentation.

\begin{table*}[tb]
    \centering
    \includegraphics[scale=1.0]{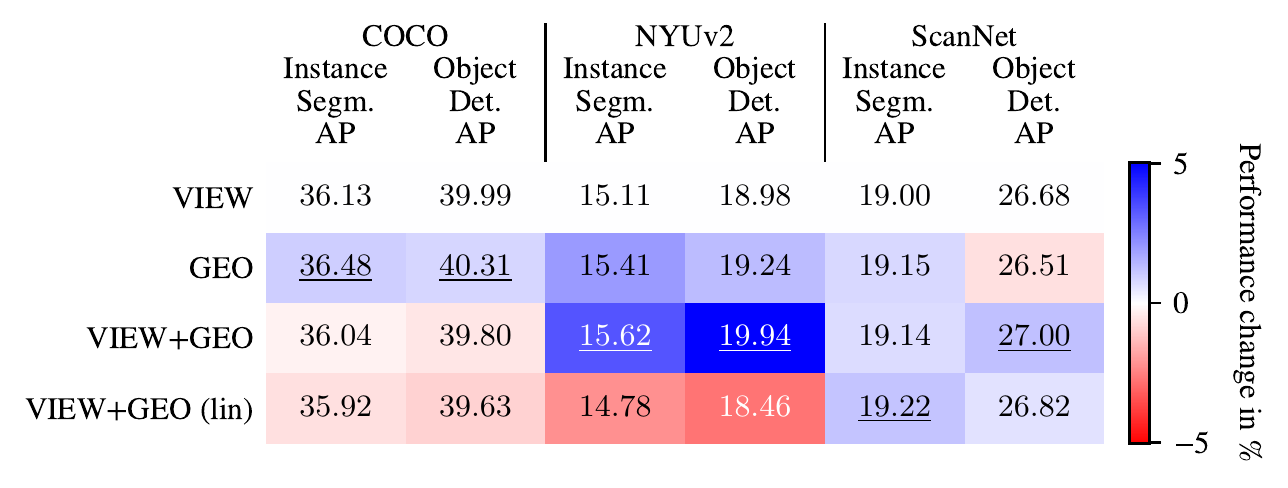}
    \caption{%
    \tr{}%
    Comparison of instance segmentation and object detection performance of the four different models (rows).
    Blue color indicates relative improvement with respect to the \mycview{} model (\protect\mabr{v}). %
    \tb{}%
    }
    \label{tab:hm_ViewvGeovViewGeo_insobj}
\end{table*}
\Cref{tab:hm_ViewvGeovViewGeo_insobj} shows that the \mycmodal{} \mabr{g} model outperforms the \mabr{v} model on 5 out of 6 task and dataset combinations.
The combined models \mabr{vgno} and \mabr{vgll} models show mixed results, perform well on \dataset{scannet}, but not on \dataset{coco}.
On \dataset{nyuv2}, their difference is most pronounced where the linear feature separation\mabr{ll} leads to the worst performance among the four model variations while omitting it results in the best performance.

Combining the losses in the \mabr{vgno} and \mabr{vgll} models seems to be beneficial within the pretraining dataset, but not when shifting domains.
Overall, although the performance differences are small, the \mabr{g} model appears to be more robust.
This supports the idea that texture-biased networks are not as robust for object-centered tasks as the depth-biased \mabr{g} network.

\tb{}
\section{Discussion}
\tr{}
The comparison on the frozen downstream tasks reveals that \mycview{} and \mycmodal{} representation alignment influence the \myprivate{} and \myshared{} information encoded in the learned representations.
While the \mabr{v} model preserves the texture and the \mabr{g} model preserves the depth, the combination \mabr{vgno} does not result in the different low-level features complementing each other, but instead a fully shared feature space is obtained where \myprivate{} information is lost and \myshared{} details are less accentuated.
\mabr{vgll} improves the combined model on all three frozen downstream tasks and raises the question of whether this means we have found a generally better representation.
The performance on the half-frozen and full finetuning downstream tasks contradict this apparent conclusion and we even find that the \mabr{g} model performs most robustly across tasks and datasets.
This shows that reducing texture-bias in networks increases robustness, similar to findings in~\cite{geirhos2019}.

On semantic segmentation, the frozen downstream task shows a distinct ranking that is persistent across datasets.
These distinct performance differences were not observed once the decoder is finetuned as well.
A possible explanation could be that the texture-bias is to a certain degree useful but also easy to recover and thus not important to learn during pretraining.
So although the strong differences on the frozen downstream tasks reveal that the diverse representations either incorporate or discard \myprivate{} information, the performance on the frozen tasks does not correlate with the finetuning downstream performance.

The comparison of \mabr{vgno} with \mabr{v} and \mabr{g} indicates different behavior on data within the same domain and across domains.
The \mabr{vgno} model excels especially in object detection, instance segmentation, and semantic segmentation on the \dataset{scannet} dataset.
A similar effect of overfitting to the pretraining dataset is also reflected in the learning rate variations shown in the supplementary material.
In a broader context, this could mean that aligning \mycmodal{} and \mycview{} representations may harm generalization across datasets but could prove especially useful when exploiting unlabelled data within the same domain for better performance on a small annotated subset.
Still, further experiments are needed to verify this hypothesis.

Our empirical study comes with certain limitations.
Although our results are obtained on various datasets and tasks, the pretraining and finetuning results are always subject to stochastic effects in the training processes.
Further, we only focused on the most obvious modality-specific information such as color and depth. %
The image reconstruction and depth prediction tasks only test some characteristics of the learned representations while other characteristics may be more indicative with respect to downstream finetuning performance.
Still, our results show that the different pretraining strategies yield representations that encode different information.

\tb{}

\section{Conclusions and future work}
\tr{}
Aligning feature representations across views and across modalities is a common approach in self-supervised learning.
Yet, the effects on the learned visual representations are often not well understood.
We showed quantitatively that, in contrast to \mycview{} representation alignment, \mycmodal{} representation alignment can lead to discarding \myprivate{} information of the individual modalities.
In the context of \mycmodal{} learning on images and point clouds, texture is considered \myprivate{} information of the visual data and depth is \myshared{} information.
Pretraining by \mycmodal{} alignment was especially useful for tasks that require a notion of the spatial layout, such as depth prediction, and also instance segmentation and object detection.
This effect is similar to the observations of previous work that shape-biased networks are more robust than texture-biased networks.
Thus, pretraining with emphasis on the \myshared{} information achieved the most robust performance when finetuning the visual backbone networks on various downstream tasks and datasets.

Merging \myprivate{} information is a fundamental goal of sensor fusion and discarding that information may harm the overall performance on such a task.
While our results have indicated that visual representations become more robust by reducing the texture-bias through \mycmodal{} alignment, future work should investigate whether these results still hold for self-supervised approaches that fuse different sensing modalities.
\tb{}

{\small
\bibliographystyle{unsrt} %
\bibliography{venues,bibfile}
}

\end{document}


\title{Supplementary material to: \\ How do Cross-View and Cross-Modal Alignment\\ Affect Representations in Contrastive Learning?}

\author{Thomas M.\ Hehn 
\and
Julian F.P.\ Kooij \\[0.5em]
Intelligent Vehicles Group \\
TU Delft, The Netherlands
\and
Dariu M.\ Gavrila
}

\maketitle
\thispagestyle{empty}

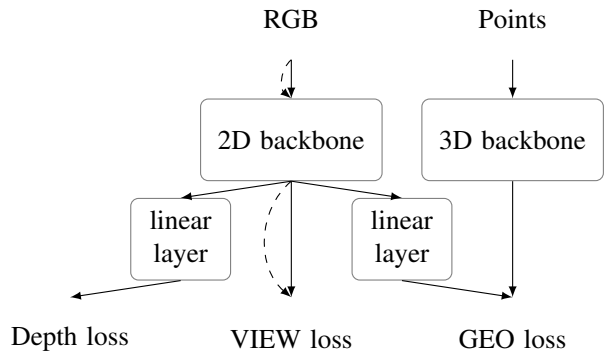
\begin{figure*}[th]
    \centering
        \newcommand{\showdepthgraph}{true} 
        \newcommand{\showcodegraph}{linear layer} 
        \resizebox{0.52\linewidth}{!}{
        \begin{tikzpicture}
    \providecommand{\showcodegraph}{false}
    \providecommand{\showdepthgraph}{false}
    \def\nodewidth{6.0em}
    \def\nodeheight{3.0em}
    \def\nodedistx{0.5}
    \tikzstyle{mainnode}=[black!50, text=black,  text centered, minimum height=\nodeheight, text width=\nodewidth]
    \tikzstyle{fixednode}=[fill=gray!25]
    \tikzstyle{NNnode}=[draw, rounded corners=0.1cm, text width=\nodewidth]
    \tikzset{>=latex}
    
    \tikzset{
    between/.style args={#1 and #2}{
         at = ($(#1)!0.5!(#2)$)
        }
    }
    
    \node[mainnode] (input_rgb) {RGB};
    \node[mainnode, right=\nodedistx of input_rgb] (input_pc) {Points};
    \node[mainnode, below=\nodedistx of input_rgb, NNnode]  (backbone2d) {2D backbone};
    \node[mainnode, below=\nodedistx of input_pc, NNnode]  (backbone3d) {3D backbone};
    
    \node[mainnode, below=(3*\nodedistx) of backbone2d]  (viewloss) {VIEW loss};
    \node[mainnode, below=(3*\nodedistx) of backbone3d]  (geoloss) {GEO loss};
    
    \path[draw,->] (input_rgb.south) -- (backbone2d.north);
    \path[draw,->] (input_pc.south) -- (backbone3d.north);
    \path[draw,->] (backbone2d.south) -- (viewloss.north);
    
    \path[->, dashed] (input_rgb.south) edge [bend right=50] (backbone2d.north);
    \path[->, dashed] (backbone2d.south) edge [bend right=50] (viewloss.north);
    
    \ifthenelse{\equal{\showcodegraph}{false}}{
        \path[draw, ->] (backbone2d.south) -- (geoloss.north);
    }{
        \node[mainnode, between=backbone2d and geoloss, NNnode, text width=(0.5 * \nodewidth)]  (w2d) {\showcodegraph};
        \path[draw, ->] (backbone2d.south) -- (w2d.north);
        \path[draw, ->] (w2d.south) -- (geoloss.north);
    }
    \ifthenelse{\equal{\showdepthgraph}{true}}{
        \node[mainnode, left=\nodedistx of viewloss]  (depthloss) {Depth loss};
        \node[mainnode, NNnode, text width=(0.5 * \nodewidth), between=backbone2d and depthloss]  (headnet) {linear\\layer};
        \path[draw, ->] (backbone2d.south) -- (headnet.north);
        \path[draw, ->] (headnet.south) -- (depthloss.north);
    }{ }
    \path[draw, ->] (backbone3d.south) -- (geoloss.north);
\end{tikzpicture}
        }
    \caption{This graph shows how the depth proxy loss is included in the Pri3D model.
    The dashed arrow indicates a second RGB view of the same scene that overlaps with the first view.
    Note that \protect\mabr{vgdll} was evaluated throughout the Pri3D paper.}
    \label{fig:pri3ddepth}
\end{figure*}

\section{Relation to results presented in Pri3D}
To the best of our knowledge, all models in Pri3D~\cite{hou2021pri3d} were pretrained with a depth loss in addition to the \mabr{v} and \mabr{g} loss.
This \textit{proxy} depth loss follows the formulation presented in~\cite{hu2019revisiting} and details are found in the official Pri3D implementation~\cite{hou2021pri3dcode}.
Note that this loss is also preceded by a linear per pixel layer that reduces the backbone output channels to a single channel (see \cref{fig:pri3ddepth}).
We did not include this proxy loss, because our goal was to focus on the effects of \mycview{} and \mycmodal{} representation alignment.
\cref{tab:hm_depthvsnodepth} show the results of the Pri3D variations once the proxy depth loss is used.
The relative performance change is given with respect to the corresponding model without the depth loss, which is not listed explicitly.
While the depth loss does have positive effects on semantic segmentation of \dataset{scannet} and \dataset{nyuv2}, it is not beneficial on the object detection and instance segmentation tasks on the same datasets.
On the other datasets and tasks it leads to mixed results.

Further, from what we understand, some models in~\cite{hou2021pri3d} were pretrained with a learning rate of 0.1 and others with 0.01.
\cref{tab:hm_learningrate} shows the effect of the learning rate change.
Within the \dataset{scannet} dataset, the higer learning rate outperforms the lower learning rate, and vice versa on the other datasets.
The pretraining learning rate thus seems to be a crucial factor that should be taken into account when evaluating models within or across datasets.

\begin{table*}[tb]
    \scalebox{0.85}{\includegraphics[]{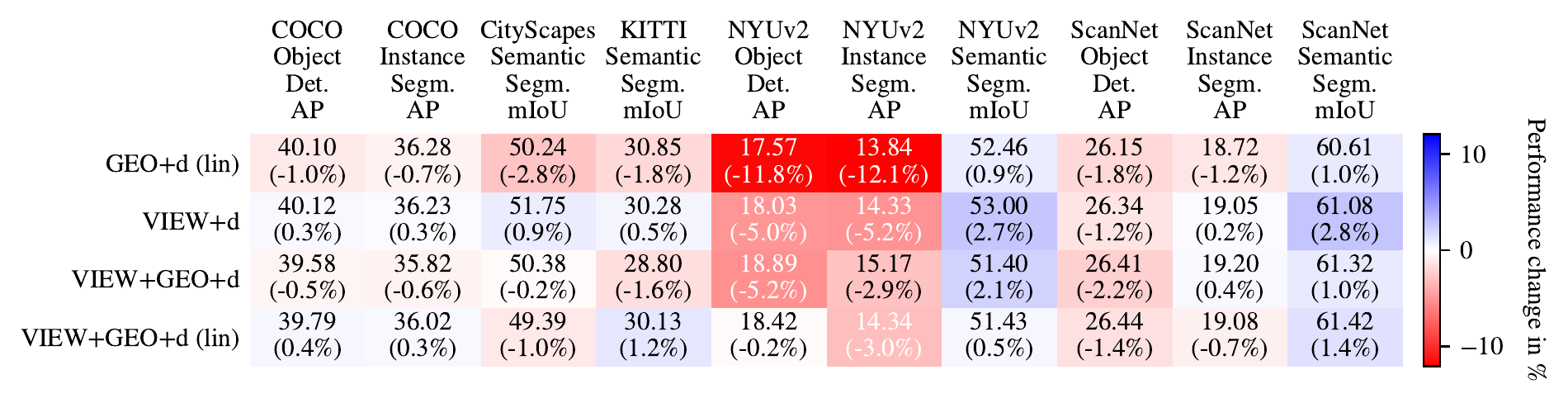}}
    \caption{Comparison of pretraining with and without the depth loss. The baseline for the performance change of per row is the same model but without the depth loss. The baseline performance is not listed explicitly here.}
    \label{tab:hm_depthvsnodepth}
\end{table*}

\begin{table*}[tb]
    \scalebox{0.85}{\includegraphics[]{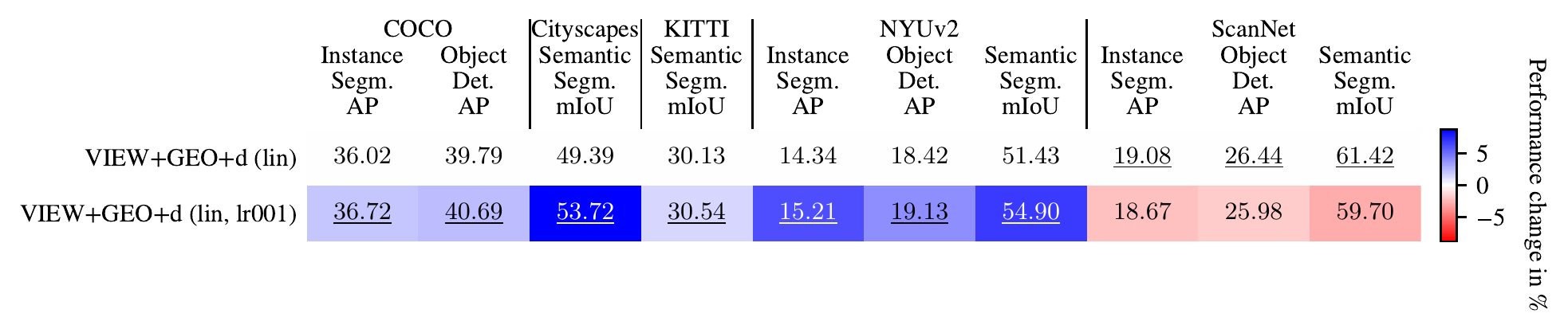}}
    \caption{\protect\mabr{vgdll} with learning rate 0.1 compared to learning rate 0.01. The relative performance is shown with respect to the baseline model \protect\mabr{vgdll}.}
    \label{tab:hm_learningrate}
\end{table*}

{\small
\bibliographystyle{unsrt} %
\bibliography{venues,bibfile}
}